\documentclass{Interspeech}
\usepackage{cite}

\interspeechcameraready

\title{One Whisper to Grade Them All}

\author{Nhan}{Phan}
\author{Anusha}{Porwal}
\author{Yaroslav}{Getman}
\author{Ekaterina}{Voskoboinik}
\author{Tamás}{Grósz}
\author{Mikko}{Kurimo}

\affiliation[nocounter]{Department of Information and Communications Engineering}{Aalto University}{Finland}
\email{\{firstname.lastname\}@aalto.fi}

\usepackage{comment}

\begin{document}

\maketitle

\begin{abstract}
    
    We present an efficient end-to-end approach for holistic Automatic Speaking Assessment (ASA) of multi-part second-language tests, developed for the 2025 Speak \& Improve Challenge. Our system's main novelty is the ability to process all four spoken responses with a single Whisper-small encoder, combine all information via a lightweight aggregator, and predict the final score. This architecture removes the need for transcription and per-part models, cuts inference time, and makes ASA practical for large-scale Computer-Assisted Language Learning systems.
    
    Our system achieved a Root Mean Squared Error (RMSE) of 0.384, outperforming the text-based baseline (0.44) while using at most 168M parameters (about 70\% of Whisper-small). Furthermore, we propose a data sampling strategy, allowing the model to train on only 44.8\% of the speakers in the corpus and still reach 0.383 RMSE, demonstrating improved performance on imbalanced classes and strong data efficiency.    

\end{abstract}

\section{Introduction}

Speaking is a core dimension of communicative competence and therefore must be assessed for any language qualification to remain valid \cite{bachman_Language_testing_fundamental1990}. Human rating of oral response, however, is time-consuming and expensive. Furthermore, rater consistency may be compromised because of fatigue \cite{ling_assessment_fatigue2014}, insufficient training \cite{david_assessment_training2015}, or limited exposure to L2 speakers’ accents \cite{park_assessment_accent2020}. Automatic Speaking Assessment (ASA) can mitigate these drawbacks by delivering scalable, objective scores while greatly reducing cost. In Computer-Assisted Language Learning (CALL) applications, ASA can also support learning by providing immediate, nonjudgmental feedback, which can lessen speaking anxiety and promote autonomous practice \cite{qiao_CALL_self_regulation}.

Early ASA systems predicted holistic proficiency from manually engineered acoustic or textual features \cite{crossley_ASA_text2013, wang_ASA2018}. 
Because these features were hand-crafted, their usefulness depended heavily on what developers chose to include, and they could overlook cues that data-driven methods might capture. Recent studies have therefore explored the use of automatically extracted features from the audio signal, allowing ASA systems to function as end-to-end graders that take only the learner’s speech as input \cite{chen_E2E_ASA2018, al-ghezi_ASA_lowresource2023, banno_ASA_wav2vec2023, banno_AssessmentL2Oral2023, qian_SpeakAmpImprove2024}. While these approaches can outperform models based on handcrafted features, they also introduce new challenges related to computational complexity. Most end-to-end models are designed to predict holistic scores for a single question or task \cite{chen_E2E_ASA2018, banno_ASA_wav2vec2023, al-ghezi_ASA_lowresource2023}; for multi-part tests, they often require multiple models or model ensembles \cite{banno_AssessmentL2Oral2023, qian_SpeakAmpImprove2024}.

\begin{figure}[t]
  \centering
  \includegraphics[width=0.7\linewidth]{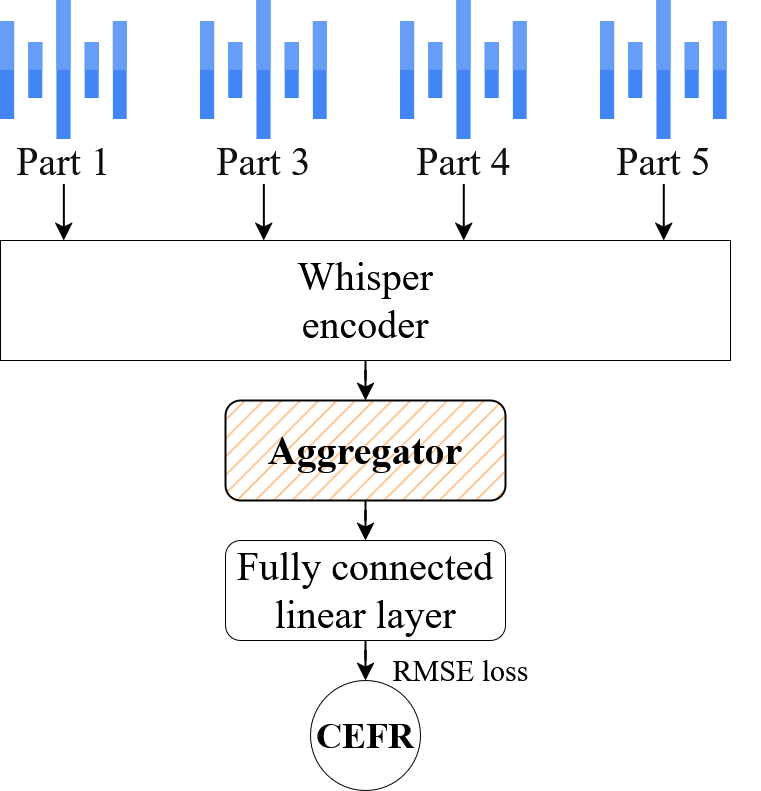}
  \caption{Overview of our model architecture.}
  \label{fig:model_overall}
\end{figure}

The requirement of multiple models to compute a holistic score for multi-part speaking tests limits the practicality of existing ASA systems, in terms of speed and computational cost. Although the inference time for each model may be similar, loading them sequentially impacts latency, while running them in parallel increases memory demands. These constraints make real-time deployment at scale more difficult. To address this, we propose a more efficient solution for CALL applications: leveraging a single Whisper small \footnote{\url{https://huggingface.co/openai/whisper-small}} encoder  \cite{radford_whisper2023} that can process all spoken responses from a multi-part test sequentially and predict the speaker’s holistic score with minimal performance trade-off (Figure~\ref{fig:model_overall}). This design reduces inference time and resource usage, making it possible to deliver immediate feedback to large numbers of learners simultaneously—a key advantage for scalable ASA systems.

We demonstrate our proposed solutions in the Speak \& Improve Challenge 2025: Spoken Language Assessment (SLA) track\cite{qian_SpeakAmpImprove2024}, where it achieved a Root Mean Squared Error (RMSE) of 0.384. This result is significantly better than the official baseline, which combined Whisper-small for automatic speech recognition (ASR) with four separate BERT-based scoring models and achieved an RMSE of 0.44.
Our architectures and detailed configurations are released as open source\footnote{\url{https://github.com/aalto-speech/slate-2025}}.

\section{Data}
\label{sec:data}

Our data comes exclusively from the Speak \& Improve Corpus 2025 (S\&I) \cite{knill_SpeakImprove2024}, as part of the SLA Challenge. The task is to develop models that assess L2 learners' language proficiency using spoken responses from parts 1, 3, 4, and 5 (part 2 is excluded). Parts 1 and 5 are short answers ($\leq$20 seconds), while parts 3 and 4 are a single long answer ($\leq$1 minute).  Each sample includes four parts: parts 1 and 5 (short responses) are first concatenated separately, and together with parts 3 and 4, form approximately four minutes of audio per speaker.

We follow the official dataset splits: 244 hours of speech for training (TRAIN), and 35 hours each for development (DEV) and evaluation (EVAL). However, we do not use the entire dataset in our experiments (see Table~\ref{tab:efficiency} for details). Each speaker's final holistic score follows the Common European Framework of Reference (CEFR) \cite{CEFR_2001}, mapped to numerical values from 2.0 (A2) to 5.5 (C1+) in 0.125-point increments. For full corpus details, see \cite{knill_SpeakImprove2024}.

One challenge in the dataset is the imbalance in the available training sample per holistic level. Here, we define edge cases as speakers with CEFR levels of A2+ or lower ($\leq 2.5$) and C1 or higher ($\geq 5.0$). While these levels account for over one-third of the possible score range, they represent only about 13\% of the data. The scarcity of edge-case samples poses an additional challenge for models like ours, which require complete responses across all parts. Moreover, prior work by Bannò et al. \cite{banno_AssessmentL2Oral2023} has shown that audio-based models may underperform on high-proficiency speakers. This further amplifies the impact of data imbalance at the C1 level and above.

\subsection{Swap Sampling Strategy}

In contrast to models trained separately for each part, our approach requires complete responses for all four parts (1, 3, 4, and 5). However, many speakers in the corpus have missing responses, reducing the usable training set to only 77.2\% of the available speaker IDs (see Table~\ref{tab:efficiency}). This limitation is particularly severe for edge cases, which are already underrepresented.

To address this, we introduce a simple but effective augmentation method called swap sampling. The core idea is to synthetically construct 4-part samples by combining responses from different speakers. For example, if speaker A is missing part 5 but speaker B has it, we can form a new full sample using parts 1, 3, and 4 from A and part 5 from B. If both A and B have all parts, we can still swap individual parts between them to generate up to 14 additional combinations. To ensure consistency, we only apply swap sampling when the score of each part is within 0.5 of the target CEFR score. This technique allows us to build larger and more balanced datasets using only the original S\&I corpus. We construct three variants of training data using this method:

\begin{itemize}
        \item Standard: 1,684 original samples with all four parts.
        \item 4x Swap Sampling (4x): expands the training set by generating approximately four times more samples via swapping. The samples were randomly chosen within 0.5 score range. 
        \item Oversamples edge scores (OE) by increasing the number of samples near the edges (CEFR $\leq$ 2.75 or $\geq$ 4.75) to 80 per score point, while reducing mid-range scores to 40. With scores ranging from 2.0 to 5.5 in 0.125 increments, this results in 1,720 samples, including both original and synthetic.
\end{itemize}

As we trained on both the TRAIN and DEV sets, we constructed a synthetic development set using swap sampling, ensuring that none of the combinations overlap with those used in training. Additionally, we create an OE$_d$ development set designed to mirror the data distribution of the OE training set. Because both OE and OE$_d$ focus on synthesized edge-case samples, they require significantly fewer speaker IDs (Table~\ref{tab:efficiency}).

\begin{table}[t]
  \caption{Unique speaker IDs used in training and development sets compared to the full S\&I corpus (TRAIN + DEV).}    
  \label{tab:efficiency}
  \centering
  \begin{tabular}{ l c c}
    \toprule
    Dataset version & Unique speaker IDs & \% of corpus \\
    \midrule    
    Standard &  1684 & 77.2\% \\    
    4x &  1871 & 85.8\% \\    
    OE & 1134 & 52.0\% \\
    OE+OE$_{d}$   & 976 &  44.8\% \\    
    \bottomrule
  \end{tabular}
  
\end{table}

\section{Experiments}

Our model architecture is based on the Whisper-small encoder. We selected an acoustic model over a text-based one, as the latter—such as BERT—requires an additional acoustic model for ASR. By using an acoustic model directly for scoring, we eliminate the need for intermediate transcription, reducing both model complexity and inference time. Among acoustic models, Whisper was chosen for its robustness in zero-shot settings \cite{radford_whisper2023}, which is particularly important when working with L2 speakers.

An overview of the architecture is in Figure~\ref{fig:model_overall}. Since Whisper can only process 30 seconds of audio at a time \cite{radford_whisper2023}, and full multi-part responses can exceed 240 seconds, our challenge lies in efficiently aggregating encoder outputs. We divide each sample into sequential, non-overlapping 30-second chunks, keeping the short chunks without masking or padding. These chunk-level representations are then passed to an Aggregator, which combines them into a fixed-size embedding used for CEFR prediction. The design of the Aggregator is critical to our model performance. While a large and complex Aggregator could retain finer details from the Whisper encoder, it would also introduce substantial computational cost. Instead, our Aggregator is lightweight and outputs a single embedding, which is passed through a fully connected layer to produce the final prediction. 

The output of the Whisper-small encoder is a tensor with two dimensions: encoder time steps ($T$) and a feature vector of size 768 for each step. To keep the model lightweight, we average over the time dimension $T$, producing a single 768-dimensional vector for each 30-second chunk. This step is distinct from the later pooling performed by the Aggregator. It compresses all frame-level information into a single embedding, trading fine-grained temporal details for computational efficiency. As a result, highly local phenomena—such as brief but severe mispronunciations—are smoothed out and may contribute less to the final representation.  

The ASA model predicts CEFR scores as continuous values ranging from 2.0 to 5.5. Although we initially experimented with Mean Squared Error (MSE) to penalise large errors more severely, in 16-bit precision MSE values can become too small and lead to gradient underflow \cite{micikevicius_mixed_fp162018}, thus we opted for Root Mean Squared Error (RMSE) as the loss function. All models were trained for 15 epochs with a learning rate of $5\times 10^{-5}$. The best-performing epoch was selected based on the synthetic development set or the OE$_d$ development set.

\subsection{Aggregator}

\begin{figure}[t]
  \centering
  \includegraphics[width=\linewidth]{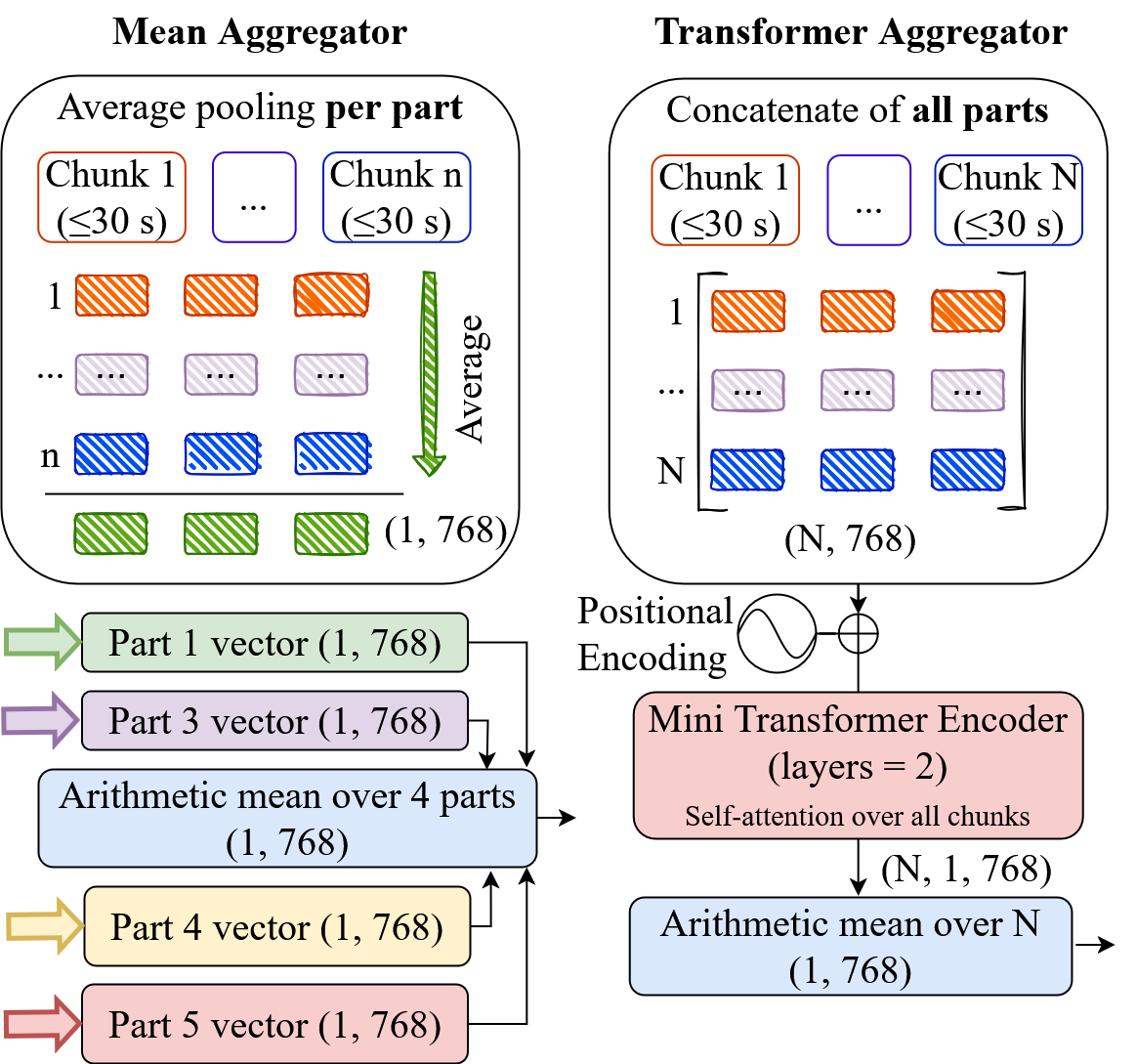}
  \caption{Schematic diagram of the two aggregation strategies. AVG averages chunk embeddings per part, while TF uses a Transformer encoder to combine all chunks.}
  \label{fig:model_aggegrator}
\end{figure}

We experimented with two aggregation strategies: \textbf{Mean Aggregator (AVG)} and \textbf{Transformer Aggregator (TF)} (Figure \ref{fig:model_aggegrator}). The first, AVG, aims for the most lightweight model by applying average pooling in two stages. Before the Aggregator, each chunk has already been time‑averaged into a 768‑dimensional vector, matching the hidden size used throughout Whisper small. AVG then averages these chunk embeddings within each part, producing one 768‑dimensional vector per part. Once all four parts (1, 3, 4, 5) are processed, AVG computes the arithmetic mean of the four part-level vectors, resulting in a single 768-dimensional vector that summarizes the entire response. This final embedding is passed through a fully connected layer to predict the CEFR score. AVG is designed to align with the scoring methodology of the corpus, which derives the final holistic score by averaging part-level scores.

The AVG aggregator is expected to lose semantic detail, as it computes a simple mean over all chunk embeddings within an answer, without preserving their order. As a result, temporal structure is discarded, and the distinct contribution of individual chunks may be diluted.  In effect, AVG primarily captures aspects of delivery (how something is said) rather than the speech content (what is said). This loss of structural information may limit the model’s ability to evaluate advanced proficiency \cite{banno_AssessmentL2Oral2023}. Notably, these higher proficiency levels also overlap with the imbalanced edge cases discussed in Section \ref{sec:data}, further degrading performance on CEFR levels of C1 or above.

To better preserve semantic and structural information, we introduce  TF aggregator, based on the encoder of the Transformer model \cite{vaswani_transformer}. To maintain efficiency, TF uses a lightweight configuration: 768-dimensional input embeddings, 4 attention heads, $4 \times 768=3072$ dimensions for the feedforward, and 2 encoder layers. The TF Aggregator is designed to retain coarse information about chunk order and cross-part dependencies, while remaining computationally efficient. Clearly 

\section{Results}

Table \ref{tab:results} compares the baseline solution's performance to our models' performance, including the best submission we made to the SLA challenge. The main evaluation metric is RMSE, but we also report RMSE$_{e}$, which measures performance on the edge cases described in Section~\ref{sec:data}. We evaluate different aggregation strategies and training set variants to test whether swap sampling improves overall performance and whether targeted oversampling helps with rare proficiency levels.

\begin{table}[t]
  \caption{Model's performance and the SLA challenge baseline. Models with 4x, OE, and OE$_d$ use the training and development configurations described in Section~\ref{sec:data}. The models chosen for our best submission are marked with *.}
  \label{tab:results}
  \centering
  \begin{tabular}{ l c c c c c}
    \toprule
    & RMSE & RMSE$_{e}$ & $\%\leq0.5$  &  $\%\leq1.0$ \\
    \midrule    
    Baseline & 0.440 & & 73.7 & 96.7 \\
    \midrule    
    AVG & 0.384 & 0.698 & 81.3 & 99.7 \\
    AVG-4x* & 0.383 & 0.717 & 81.3 & 99.3 \\
    AVG-OE & 0.387 & 0.613 & 81.3 & 98.7 \\
    AVG-OE+OE$_d$ & 0.386 & 0.618 & 80.3 & 99.7 \\
    TF & \textbf{0.372} & 0.637 & \textbf{81.7} & 99.7 \\
    TF-4x & 0.384 & 0.675 & 79.7 & 99.0 \\
    TF-OE* & 0.388 & 0.626 & 79.0 & 98.7 \\
    TF-OE+OE$_d$ & 0.383 & \textbf{0.597} & 79.3 & 99.3 \\
    \midrule
    Submission & 0.384 & 0.716 & 81.3 & 99.3 \\
    \bottomrule
  \end{tabular}
  
\end{table}

From the 4x model performance, we observe that increasing the number of swap samples does not improve RMSE metric, which remains around 0.383. We hypothesise this is because most added samples are synthetic combinations without novel information, and only a few represent edge cases. Furthermore, the synthetic development set did not consistently help select the best-performing checkpoints, as shown in Figure~\ref{fig:reliability}. 

In contrast, the oversample edge (OE) models consistently perform better on edge cases across both aggregators. This suggests that targeted oversampling can partially mitigate the effects of data imbalance, in line with prior findings \cite{rathpisey_sampling2019, lun_oversampling2024}. Notably, the OE models validated with the oversample edge development set (OE+OE$_d$) also achieve comparable overall performance to other configurations, despite using significantly fewer speaker IDs. This result highlights their advantage in terms of data efficiency, which we discuss further in Section~\ref{sec:efficiency}.

While our best-performing individual model was TF with an RMSE of 0.372, it was not used in our official SLA submission. During development, we found that TF models exhibited greater performance fluctuations across training epochs - occasionally achieving lower RMSE, but also prone to sharp degradation depending on checkpoint selection (see Figure~\ref{fig:reliability}). In contrast, AVG models were more stable: due to their averaging design, they consistently produced similar RMSE scores around 0.383 regardless of the chosen epoch. This stability is particularly advantageous in evaluation settings where access to the test set is restricted, as was the case with the SLA challenge.

As discussed in Section~\ref{sec:data}, TF models tend to outperform AVG models on edge cases, achieving lower RMSE$_e$ across training configurations. To leverage this, we implemented a model combination strategy inspired by Mixture of Experts \cite{masoudnia_mixture_MoE_2014}, in which the TF model handled predictions for edge-case samples while AVG handled the rest. Our submitted model combined AVG-4x (for overall stability) and TF-OE (for better edge-case prediction). However, improved edge-case performance does not necessarily translate to better overall scores, as the official test set contained only 21 edge samples (7\%). Furthermore, our TF-OE model, despite performing better in edge cases, still predicted values that fell outside the edge-score range (e.g., predicting 4.9 for a true score of 5.5). These errors resulted in those predictions being excluded by the selection mechanism in our Mixture of Experts scheme, ultimately limiting the impact of TF-OE in the final submission.

\subsection{Data and Computational Efficiency}
\label{sec:efficiency}

We evaluate efficiency from two perspectives: data usage and computational cost. First, in terms of data efficiency, OE+OE$_d$ models use significantly fewer samples than other configurations—only 44.8\% of speaker IDs from the official TRAIN and DEV sets (Table~\ref{tab:efficiency})—while still achieving competitive performance, particularly on edge cases. For reference, due to architectural constraints, even our standard model only uses 77.2\% of the corpus, as it requires complete responses for all four parts. These results raise a broader question: could future systems benefit more from selectively collecting underrepresented edge cases, rather than expanding common samples that can already be simulated through different sampling techniques?

Second, in terms of computational efficiency, our model architecture is designed for low-cost inference. The baseline system relies on Whisper for transcription and four separate BERT models for scoring each part, followed by score averaging. In contrast, our system uses only a single Whisper encoder and a lightweight Aggregator to directly predict the holistic score, skipping both the Whisper decoder and ASR entirely.

On a standard CPU setup (Intel Xeon E5-2680 v3 2.50 GHz, 4 GB RAM, no GPU), our inference time is about 60 seconds for a 4-part response totalling 240 seconds of audio. This demonstrates that our model can serve as a practical, low-cost solution for real-time or large-scale CALL systems. With a GPU, feedback can be delivered almost instantly; for example, inference on an RTX 4070 takes less than one second.

In terms of model size, our AVG configuration uses 154M parameters, while the TF model uses 168M, both substantially smaller than the full Whisper-small model (244M). Although our models currently load the Whisper decoder (88M) for future research (see Section~\ref{sec:validity}), it is not used during inference.

\subsection{Reliability and Validity in ASA}
\label{sec:validity}

\begin{figure}[t]
  \centering
  \includegraphics[width=0.9\linewidth]{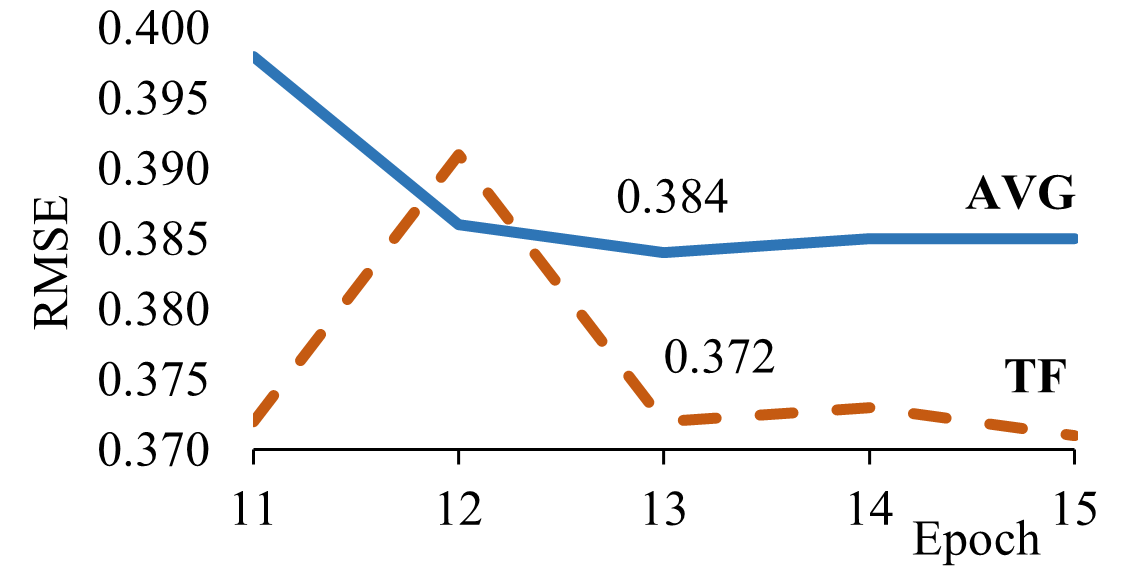}
  \caption{RMSE on the EVAL set across training epochs for AVG and TF. Based on development set performance, epoch 13 was selected as the best checkpoint.
  }
  \label{fig:reliability}
\end{figure}

In language testing, the quality of a speaking assessment is typically evaluated through three key principles: reliability, validity, and practicality \cite[pp.175–186]{luomaAssessingSpeaking2004}. Although these criteria were originally developed for human scoring, they offer a useful framework for evaluating ASA systems. Our architecture addresses practicality through model simplicity and low inference cost. Here we focus on (i) reliability, interpreted as score stability, and (ii) validity, assessed here by a single criterion: the model’s reaction to missing or mismatched content \cite{cheng_OfftopicASA2011, yoon_contentmodelingASA2019}, which offers a useful yet partial indicator of construct coverage.

\textbf{Reliability.} Figure~\ref{fig:reliability} plots RMSE on the EVAL set across training epochs. AVG converges after epoch 12 with minor variation, whereas TF shows larger fluctuations, including a sharp spike at the same epoch. This suggests that AVG is more robust to training fluctuations, which is especially important in blind evaluation settings where tuning on the test set is not possible. AVG's double mean pooling, which smooths out training noise, likely contributes to this consistency,


    
  

\begin{table}[t]
    \caption{Change in predicted score ($\Delta$) after inserting dummy or swapped speech (higher = greater content
           sensitivity).}
  \centering
  \label{tab:validity}
  \centering
  \begin{tabular}{ l r r r}
    \toprule
    & $\Delta \uparrow$ & $\%\geq0.25 \uparrow$  &  $\%\geq0.5 \uparrow$ \\   
    \midrule    
    \multicolumn{4}{c}{Dummy speech} \\
    \midrule    
    AVG                 &  0.120 & 16.7 & 1.0 \\    
    TF                  &  \textbf{0.308} & \textbf{63.0} & \textbf{10.3} \\    
    \midrule    
    \multicolumn{4}{c}{Part swapping} \\
    \midrule    
    AVG                 &  0.000 & 0.0 & 0.0 \\    
    TF                  &  -0.012 & 0.0 & 0.0 \\    
    \bottomrule
  \end{tabular}
  
\end{table}

\textbf{Validity.} We designed two controlled tests to evaluate how sensitive the models are to speech content, all of which were done on the EVAL set. In the first test (dummy speech), we replaced one part with 60 seconds of silence. In the second (part swapping), we swapped answers between parts 1 and 3, and between parts 4 and 5, simulating off-topic answers. For each test, we computed $\Delta=\hat y_{\text{orig}}-\hat y_{\text{edit}}$, the difference in predicted score of the original input compared to the edited input. Table~\ref{tab:validity} reports the percentage of samples with $\Delta \geq 0.25$ and $\Delta \geq 0.5$.

The dummy speech test shows that TF is significantly more sensitive to missing content. When all four parts are silent, TF predicts a CEFR score of 2.02 (A2 - lowest score possible for this dataset), while AVG still predicts 3.16. However, both models are insensitive to part swapping, confirming previous findings that ASA systems based solely on acoustic features primarily evaluate speech delivery rather than speech content \cite{banno_AssessmentL2Oral2023, xi_SpeechRater2008}.

These findings highlight a critical limitation: ASA systems can be easily misled by fluent but off-topic or irrelevant content. Despite growing interest in acoustic-based ASA, this gap in validity remains underexplored. For high-stakes use cases, we recommend pairing the acoustic ASA with a dedicated content assessment module, such as the one proposed in \cite{phan24_interspeech}. In low-stakes CALL settings, we had initially planned to use the Whisper decoder and BERT to condition the grader on the question. However, due to a technical issue, access to task questions was not available during the SLA challenge. 


\section{Conclusion}

In this paper, we present a simple architecture that uses a single Whisper encoder to predict holistic speaking scores from multi-part responses, eliminating the need for separate per-part models.  Our system outperforms the SLA challenge baseline, which uses Whisper small and four BERT-based models. Combined with our swap oversampling technique, the model demonstrates strong data efficiency: it uses less than half of the available speakers from the S\&I corpus while maintaining competitive performance, making it a promising solution for imbalanced data. 

However, like other acoustic-only ASA systems, our model remains insensitive to speech content. Future work should explore integrating content-aware components, enabling the model to assess both what is said and how it is said, thereby improving validity and feedback quality in CALL applications.

\section{Acknowledgments}
This work was partly funded by the Research Council of Finland's grants 322625, 345790, 355587 and 365233. The computational resources were provided by the Aalto Science-IT.

\bibliographystyle{IEEEtran}
\bibliography{zotero}

\begin{thebibliography}{10}
\providecommand{\url}[1]{#1}
\csname url@samestyle\endcsname
\providecommand{\newblock}{\relax}
\providecommand{\bibinfo}[2]{#2}
\providecommand{\BIBentrySTDinterwordspacing}{\spaceskip=0pt\relax}
\providecommand{\BIBentryALTinterwordstretchfactor}{4}
\providecommand{\BIBentryALTinterwordspacing}{\spaceskip=\fontdimen2\font plus
\BIBentryALTinterwordstretchfactor\fontdimen3\font minus \fontdimen4\font\relax}
\providecommand{\BIBforeignlanguage}[2]{{%
\expandafter\ifx\csname l@#1\endcsname\relax
\typeout{** WARNING: IEEEtran.bst: No hyphenation pattern has been}%
\typeout{** loaded for the language `#1'. Using the pattern for}%
\typeout{** the default language instead.}%
\else
\language=\csname l@#1\endcsname
\fi
#2}}
\providecommand{\BIBdecl}{\relax}
\BIBdecl

\bibitem{bachman_Language_testing_fundamental1990}
L.~F. Bachman, \emph{Fundamental considerations in language testing}.\hskip 1em plus 0.5em minus 0.4em\relax Oxford university press, 1990.

\bibitem{ling_assessment_fatigue2014}
G.~Ling, P.~Mollaun, and X.~Xi, ``A study on the impact of fatigue on human raters when scoring speaking responses,'' \emph{Language Testing}, vol.~31, no.~4, pp. 479--499, 2014.

\bibitem{david_assessment_training2015}
L.~Davis, ``The influence of training and experience on rater performance in scoring spoken language,'' \emph{Language Testing}, vol.~33, no.~1, pp. 117--135, 2016.

\bibitem{park_assessment_accent2020}
M.~S. Park, ``Rater effects on {{L2}} oral assessment: {{Focusing}} on accent familiarity of {{L2}} teachers,'' \emph{Language Assessment Quarterly}, vol.~17, no.~3, pp. 231--243, 2020.

\bibitem{qiao_CALL_self_regulation}
H.~Qiao and A.~Zhao, ``Artificial intelligence-based language learning: Illuminating the impact on speaking skills and self-regulation in {{Chinese EFL}} context,'' \emph{Frontiers in Psychology}, vol. Volume 14 - 2023, 2023.

\bibitem{crossley_ASA_text2013}
S.~Crossley and D.~McNamara, ``Applications of text analysis tools for spoken response grading.'' \emph{Language Learning \& Technology}, vol.~17, no.~2, pp. 171--192, 2013.

\bibitem{wang_ASA2018}
Y.~Wang, M.~Gales, K.~Knill, K.~Kyriakopoulos, A.~Malinin, R.~Van~Dalen, and M.~Rashid, ``Towards automatic assessment of spontaneous spoken {{English}},'' \emph{Speech Communication}, vol. 104, pp. 47--56, Nov. 2018.

\bibitem{chen_E2E_ASA2018}
L.~Chen, J.~Tao, S.~Ghaffarzadegan, and Y.~Qian, ``End-to-{{End Neural Network Based Automated Speech Scoring}},'' in \emph{2018 {{IEEE International Conference}} on {{Acoustics}}, {{Speech}} and {{Signal Processing}} ({{ICASSP}})}.\hskip 1em plus 0.5em minus 0.4em\relax Calgary, AB: IEEE, Apr. 2018, pp. 6234--6238.

\bibitem{al-ghezi_ASA_lowresource2023}
R.~Al-Ghezi, Y.~Getman, E.~Voskoboinik, M.~Singh, and M.~Kurimo, ``Automatic {Rating} of {Spontaneous} {Speech} for {Low}-{Resource} {Languages},'' in \emph{2022 IEEE Spoken Language Technology Workshop (SLT)}, 2023, pp. 339--345.

\bibitem{banno_ASA_wav2vec2023}
S.~Bannò and M.~Matassoni, ``{Proficiency Assessment of L2 Spoken English Using Wav2Vec 2.0},'' in \emph{2022 IEEE Spoken Language Technology Workshop (SLT)}, 2023, pp. 1088--1095.

\bibitem{banno_AssessmentL2Oral2023}
S.~Bann{\`o}, K.~M. Knill, M.~Matassoni, V.~Raina, and M.~Gales, ``Assessment of {{L2 Oral Proficiency Using Self-Supervised Speech Representation Learning}},'' in \emph{9th {{Workshop}} on {{Speech}} and {{Language Technology}} in {{Education}} ({{SLaTE}})}.\hskip 1em plus 0.5em minus 0.4em\relax ISCA, Aug. 2023, pp. 126--130.

\bibitem{qian_SpeakAmpImprove2024}
\BIBentryALTinterwordspacing
M.~Qian, K.~Knill, S.~Banno, S.~Tang, P.~Karanasou, M.~J.~F. Gales, and D.~Nicholls, ``Speak \& {Improve} {Challenge} 2025: {Tasks} and {Baseline} {Systems},'' 2024. [Online]. Available: \url{https://arxiv.org/abs/2412.11985}
\BIBentrySTDinterwordspacing

\bibitem{radford_whisper2023}
A.~Radford, J.~W. Kim, T.~Xu, G.~Brockman, C.~Mcleavey, and I.~Sutskever, ``{Robust Speech Recognition via Large-Scale Weak Supervision},'' in \emph{Proceedings of the 40th International Conference on Machine Learning}, vol. 202.\hskip 1em plus 0.5em minus 0.4em\relax PMLR, 2023, pp. 28\,492--28\,518.

\bibitem{knill_SpeakImprove2024}
\BIBentryALTinterwordspacing
K.~Knill, D.~Nicholls, M.~Gales, M.~Qian, and P.~Stroinski, ``\BIBforeignlanguage{en}{Speak \& {Improve} {Corpus} 2025: an {L2} {English} {Speech} {Corpus} for {Language} {Assessment} and {Feedback}},'' Apollo - University of Cambridge Repository, Tech. Rep., Dec. 2024. [Online]. Available: \url{https://www.repository.cam.ac.uk/handle/1810/377542}
\BIBentrySTDinterwordspacing

\bibitem{CEFR_2001}
\BIBentryALTinterwordspacing
{Council of Europe}, \emph{Common {{European Framework}} of {{Reference}} for {{Languages}}: Learning, Teaching, Assessment}.\hskip 1em plus 0.5em minus 0.4em\relax Cambridge University Press, 2001. [Online]. Available: \url{https://rm.coe.int/1680459f97}
\BIBentrySTDinterwordspacing

\bibitem{micikevicius_mixed_fp162018}
P.~Micikevicius, S.~Narang, J.~Alben, G.~Diamos, E.~Elsen, D.~Garcia, B.~Ginsburg, M.~Houston, O.~Kuchaiev, G.~Venkatesh, and H.~Wu, ``Mixed precision training,'' in \emph{International Conference on Learning Representations}, 2018.

\bibitem{vaswani_transformer}
A.~Vaswani, N.~Shazeer, N.~Parmar, J.~Uszkoreit, L.~Jones, A.~N. Gomez, {\L}.~Kaiser, and I.~Polosukhin, ``Attention is all you need,'' \emph{Advances in neural information processing systems}, vol.~30, 2017.

\bibitem{rathpisey_sampling2019}
H.~Rathpisey and T.~B. Adji, ``Handling {{Imbalance Issue}} in {{Hate Speech Classification}} using {{Sampling-based Methods}},'' in \emph{2019 5th {{International Conference}} on {{Science}} in {{Information Technology}} ({{ICSITech}})}.\hskip 1em plus 0.5em minus 0.4em\relax Yogyakarta, Indonesia: IEEE, Oct. 2019, pp. 193--198.

\bibitem{lun_oversampling2024}
T.~M. Lun, E.~Voskoboinik, R.~Al-Ghezi, T.~Grosz, and M.~Kurimo, ``Oversampling, augmentation and curriculum learning for speaking assessment with limited training data,'' in \emph{Interspeech 2024}, 2024, pp. 4019--4023.

\bibitem{masoudnia_mixture_MoE_2014}
S.~Masoudnia and R.~Ebrahimpour, ``Mixture of experts: a literature survey,'' \emph{Artificial Intelligence Review}, vol.~42, pp. 275--293, 2014.

\bibitem{luomaAssessingSpeaking2004}
S.~Luoma, \emph{Assessing speaking}.\hskip 1em plus 0.5em minus 0.4em\relax Cambridge University Press, 2004.

\bibitem{cheng_OfftopicASA2011}
J.~Cheng and J.~Shen, ``Off-topic detection in automated speech assessment applications,'' in \emph{Interspeech 2011}.\hskip 1em plus 0.5em minus 0.4em\relax ISCA, Aug. 2011, pp. 1597--1600.

\bibitem{yoon_contentmodelingASA2019}
S.-Y. Yoon and C.~Lee, ``Content modeling for automated oral proficiency scoring system,'' in \emph{Proceedings of the fourteenth workshop on innovative use of {NLP} for building educational applications}, 2019, pp. 394--401.

\bibitem{xi_SpeechRater2008}
X.~Xi, D.~Higgins, K.~Zechner, and D.~M. Williamson, ``Automated scoring of spontaneous speech using {{SpeechRater}} v1.0,'' \emph{ETS Research Report Series}, vol. 2008, no.~2, pp. i--102, 2008.

\bibitem{phan24_interspeech}
N.~Phan, A.~{von Zansen}, M.~Kautonen, E.~Voskoboinik, T.~Grosz, R.~Hilden, and M.~Kurimo, ``Automated content assessment and feedback for {{Finnish L2}} learners in a picture description speaking task,'' in \emph{Interspeech 2024}, 2024, pp. 317--321.

\end{thebibliography}

\end{document}